# Pseudo Fuzzy Set

Sukanta Nayak and Snehashish Chakraverty

Department of Mathematics, National Institute of Technology Rourkela,

Odisha, India-769008

Abstract

Here a novel idea to handle imprecise or vague set viz. Pseudo fuzzy set has been proposed. Pseudo fuzzy set is a triplet of element and its two membership functions. Both the membership functions may or may not be dependent. The hypothesis is that every positive sense has some negative sense. So, one membership function has been considered as positive and another as negative. Considering this concept, here the development of Pseudo fuzzy set and its property along with Pseudo fuzzy numbers has been discussed.

1. **Introduction**

In real world problem every system contains uncertainty which may not be avoided. Traditionally these uncertainties have been handled by using probabilistic approach. Due to the involvement of linguistic variables and imprecise parameters in the system, it may be very difficult to get large number of experimental or observed data. In this context, a great philosopher (Zadeh, 1965) introduced the concept of fuzzy which may help to manage uncertainty and reduce the difficulty of availability of large number of data. Further the fuzzy set theory has been extensively used by various authors viz.(Chakraverty S. and Nayak S., 2013; Dubois D., 1980; Hanss Michael, 2005; Klir, 1997; Nayak S. and Chakraverty S., 2013; Zimmermann HJ., 1991). Recently, (Nayak S. and Chakraverty S., 2013) has developed a transformation method and modified fuzzy numbers into crisp form. This transformation has been used in various problems and the uncertain solution has been investigated.

From the above literature it has been revealed that only one membership function has been used for the each element and there may be some extension. The Newtonian philosophy viz. "Every action has equal and opposite reaction" may give us some idea about the involvement of some other membership function in addition with the membership function used in fuzzy set. The law of nature shows that if there is positive then by default some negative. So, in view of this we have introduced another membership function and the modified fuzzy number is named as Pseudo fuzzy set.

The above concept of pseudo fuzzy set has been systematically presented in the following sections. In the second section the development of Pseudo fuzzy and various properties have been discussed. Further in third section Pseudo fuzzy number have also been introduced.

## 2. Pseudo Fuzzy set and its property

Uncertainty involve in almost every system which may be vague or imprecise value. In real world problem, every system has uncertainty. The uncertainty may be probabilistic or non-probabilistic (i.e. imprecise or vague). These imprecise or vague uncertainties are considered as fuzzy, where a fuzzy set is a pair of element and its membership function. The fuzzy set $\tilde{X}$ may be defined as $\tilde{X} = \{(x, \mu_x) \mid x \in \Re, \mu_x \in [0, 1]\}$. But as we see in more details, we find that there involve another membership function $\lambda_x : X \to [-1, 0]$.

Uncertainty may be divided into negative and positive. Nothing is perfect. If we consider the effect as positive then the cause may be assumed as negative is positive then by default there is some negative which may balance or unbalance the system that may not be avoided. If positive effect has membership function $\mu_x : X \to [0, 1]$ then negative effect (cause) has membership function $\lambda_x : X \to [-1, 0]$.

If we study only positive part then we may not define the system. We have to study the negative impact.

In view of the above we may now define a system as a triplet ($x$, $\mu_x$, $\lambda_x$) where

$$\tilde{X} = \{x \in \Re \mid \mu_x \in [0, 1], \lambda_x \in [-1, 0]\}$$

In this regard some property may be written as

Property

i) $\mu_x \in [0, 1], \quad \lambda_x \in [0, 1]$

ii) $|\mu_x| \leq 1, \quad |\lambda_x| \leq 1$

iii) $0 \leq |\mu_x| + |\lambda_x| \leq 2$

Accordingly we may define now following three cases

a) $0 \leq |\mu_x| + |\lambda_x| \leq 1$

b) $|\mu_x| + |\lambda_x| = 1$

c) $1 \leq |\mu_x| + |\lambda_x| \leq 2$

Example

Cold Fever

When a person is affected by cold fever, then we may see that temperature of the body rises as well as body feels cold inside. Further it may be seen from the case study that when the body temperature rises i.e. for high fever we feel colder for this type of cold fever. So from this we may see that the hotness and coldness occurs simultaneously and so membership functions should have been defined for hotness and coldness accordingly.

Here, $\mu$ correspond the hotness of the body and $\lambda$ correspond the coldness felt by the body, where $\mu \in [0,1]$ and $\lambda \in [-1,0]$.

## 3. Pseudo fuzzy number

Depending upon the nature of the uncertainty Pseudo Fuzzy Set may be classified as follows.

(i) <u>Dependent Pseudo Fuzzy Set</u>

In this case, both the membership functions are complement to each other. Let us consider $x \in X$ be an arbitrary element of pseudo fuzzy set $X$ and $\mu$ and $\lambda$ are the membership functions then $|\mu_x| + |\lambda_x| = 1$.

The dependent Pseudo Triangular Fuzzy Number (PTFN) may be represented as follows

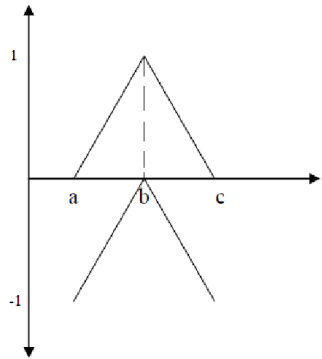

Fig. 1. Pseudo TFN (Dependent)

$$\mu_x = \begin{cases} 0, & x \le a \\ \dfrac{x-a}{b-a}, & a \le x \le b \\ \dfrac{c-x}{c-b}, & b \le x \le c \\ 0, & x \ge c \end{cases} \qquad \lambda_x = \begin{cases} -1, & x \le a \\ \dfrac{x-b}{b-a}, & a \le x \le b \\ \dfrac{b-x}{c-b}, & b \le x \le c \\ -1, & x \ge c \end{cases}$$

(ii) <u>Independent Pseudo Fuzzy Set</u>

In this case, both the membership functions are complement to each other. Let us consider $x \in X$ be an arbitrary element of pseudo fuzzy set $X$ and $\mu$ and $\lambda$ are the membership functions then $0 \le |\mu_x| + |\lambda_x| \le 1$ or $0 \le |\mu_x| + |\lambda_x| \le 2$.

The independent Pseudo Triangular Fuzzy Number (PTFN) may be represented as follows

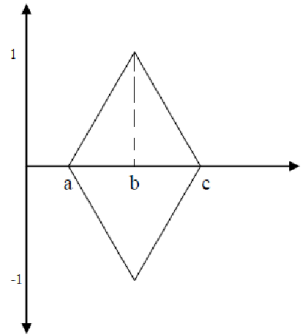

Fig. 2. Pseudo TFN (Independent)

$$\mu_x = \begin{cases} 0, & x \le a \\ \dfrac{x-a}{b-a}, & a \le x \le b \\ \dfrac{c-x}{c-b}, & b \le x \le c \\ 0, & x \ge c \end{cases} \qquad \lambda_x = \begin{cases} 0, & x \le a \\ \dfrac{a-x}{b-a}, & a \le x \le b \\ \dfrac{x-c}{c-b}, & b \le x \le c \\ 0, & x \ge c \end{cases}$$